\documentclass[11pt]{article}

\usepackage[margin=1in]{geometry}
\usepackage[utf8]{inputenc}
\usepackage[T1]{fontenc}
\usepackage{hyperref}
\usepackage{url}
\usepackage{booktabs}
\usepackage{amsfonts}
\usepackage{amsmath}
\usepackage{nicefrac}
\usepackage{microtype}
\usepackage{xcolor}
\usepackage{graphicx}
\usepackage{subcaption}

\title{The Instability of Safety: How Random Seeds and Temperature Expose Inconsistent LLM Refusal Behavior}

\author{%
  Erik Larsen \\
  \texttt{elarsen.mailbox@gmail.com} \\
}

\begin{document}

\maketitle

\begin{abstract}
Current safety evaluations of large language models rely on single-shot testing, implicitly assuming that model responses are deterministic and representative of the model's safety alignment. We challenge this assumption by investigating the stability of safety refusal decisions across random seeds and temperature settings. Testing four instruction-tuned models from three families (Llama 3.1 8B, Qwen 2.5 7B, Qwen 3 8B, Gemma 3 12B) on 876 harmful prompts across 20 different sampling configurations (4 temperatures $\times$ 5 random seeds), we find that \textbf{18--28\% of prompts exhibit decision flips} (the REFUSE/PARTIAL/COMPLY label changes across configurations), depending on the model. Our Safety Stability Index (SSI) reveals that higher temperatures significantly reduce decision stability (Friedman $\chi^2 = 396.81$, $p < 0.001$), with mean within-temperature SSI dropping from 0.977 at temperature 0.0 to 0.942 at temperature 1.0. We validate our findings across all model families using Claude 3.5 Haiku as a unified external judge, achieving 89.1\% inter-judge agreement with the Llama 70B judge on the two models both judges labeled (Cohen's $\kappa = 0.62$). Within each model, prompts with higher compliance rates exhibit lower stability (Spearman $\rho = -0.47$ to $-0.70$, all $p < 0.001$), indicating that models ``waver'' more on borderline requests. These findings demonstrate that single-shot safety evaluations are insufficient for reliable safety assessment and that evaluation protocols must account for stochastic variation in model behavior. We show that, for Llama 3.1 8B, single-shot evaluation agrees with multi-sample ground truth only 92.5\% of the time when pooling across temperatures (98.7\% at greedy decoding down to 90.3\% at temperature 1.0), and recommend scaling samples with temperature---one at greedy, three at low temperature, more at higher temperatures (where three reach only $\sim$95\%), and ten when pooling---rather than a single flat threshold, for reliable safety assessment.
\end{abstract}

\section{Introduction}

As large language models (LLMs) are increasingly deployed in real-world applications, ensuring their safety has become paramount. Current safety evaluation methodologies typically assess model responses to harmful prompts through single-shot testing: each prompt is evaluated once, and the model's response is classified as either refusing or complying with the harmful request. This approach implicitly assumes that model responses are deterministic and that a single sample accurately represents the model's safety behavior.

However, modern LLMs employ stochastic sampling during inference, introducing variability through both temperature-controlled randomness and random seed initialization. While this variability is well-documented for general generation tasks, its impact on safety-critical decisions remains underexplored. If safety decisions are unstable across sampling configurations, single-shot evaluations may significantly misrepresent a model's true safety profile---either overestimating safety by catching a safe sample from an unstable prompt, or underestimating it by observing a failure case that rarely occurs.

In this work, we systematically investigate the stability of LLM safety refusal behavior by testing the same harmful prompts across multiple random seeds and temperature settings. We introduce the \textbf{Safety Stability Index (SSI)}, a metric that quantifies how consistently a model makes the same safety decision across different sampling configurations. We evaluate four instruction-tuned models from three families---Llama 3.1 8B Instruct, Qwen 2.5 7B Instruct, Qwen 3 8B, and Gemma 3 12B Instruct---on 876 harmful prompts (491 from AdvBench and 385 from HarmBench) across 20 configurations (4 temperatures $\times$ 5 random seeds), generating 70,080 total responses. To validate our findings and address potential same-family judge bias, we additionally evaluate the newer models using Claude 3.5 Haiku as an external judge.

Our findings reveal significant instability in safety decisions:
\begin{itemize}
    \item \textbf{24.9\% of prompts exhibit decision flips} across sampling configurations on average, ranging from 18--28\% depending on the model
    \item \textbf{Temperature significantly affects stability} (Friedman $\chi^2 = 396.81$, $p < 0.001$): Lower temperatures yield more stable decisions (mean within-temperature SSI = 0.977 at temp 0.0) compared to higher temperatures (mean SSI = 0.942 at temp 1.0)
    \item \textbf{Instability is consistent across model families}: All four models from three families (Llama, Qwen, Gemma) exhibit instability patterns despite different training approaches, suggesting this may be a common property of current safety training in small-to-medium scale models (7B--12B parameters)
    \item \textbf{Borderline prompts exist}: 7.9\% of prompts fall in the highly unstable range (SSI 0.4--0.6), predominantly copyright-related requests where the model is uncertain whether to comply
    \item \textbf{Single-shot evaluation is unreliable, and the samples needed scale with temperature}: For Llama 3.1 8B, a single sample agrees with ground truth only 92.5\% of the time when pooling across temperatures (98.7\% at greedy decoding down to 90.3\% at t=1.0); reaching $>$97\% agreement takes one sample at t=0.0, three at low temperature, more at higher temperatures (where three reach only $\sim$95\%), and ten when pooling across temperatures
\end{itemize}

These results have important implications for safety evaluation practices. Single-shot testing may give a false sense of security or unnecessarily penalize models depending on which random seed is used. We argue that safety benchmarks should report stability metrics alongside accuracy, and that deployment configurations should carefully consider temperature settings to maximize both safety and consistency.

\section{Related Work}

\textbf{Safety Evaluation Benchmarks.} Safety evaluation of large language models has become a critical research area, with several benchmarks developed to assess model behavior on harmful prompts. BeaverTails~\cite{ji2023beavertails} provides a human-preference dataset for safety alignment covering diverse harm categories. HarmBench~\cite{mazeika2024harmbench} offers a standardized framework for automated red teaming and robust refusal evaluation. AdvBench~\cite{zou2023advbench} focuses on adversarial attacks against aligned models. These benchmarks typically rely on single-shot testing---each prompt is evaluated once with a fixed random seed and temperature---implicitly assuming that model responses are deterministic or that a single sample adequately represents the model's safety behavior.

\textbf{LLM Calibration and Consistency.} Recent work has investigated whether language models can accurately assess their own uncertainty. Kadavath et al.~\cite{kadavath2022language} showed that models can estimate the probability of their answers being correct. Huang et al.~\cite{huang2024survey} provide a comprehensive survey of LLM calibration, highlighting the gap between model confidence and actual accuracy. Lin et al.~\cite{lin2024generating} study methods for eliciting confidence from LLMs. However, this work focuses on factual accuracy rather than safety decisions, leaving open the question of whether safety refusal behavior is similarly well-calibrated.

\textbf{Temperature Effects on Generation.} The temperature parameter controls the randomness of LLM sampling by scaling the logit distribution before sampling~\cite{holtzman2020curious}. Higher temperatures increase diversity but can reduce coherence. Renze and Guven~\cite{renze2024temperature} study how temperature affects problem-solving performance. While temperature effects on general generation quality are well-documented, no prior work has systematically quantified how temperature affects the consistency of safety-critical decisions.

\textbf{Red Teaming and Jailbreaking.} Automated red teaming techniques~\cite{perez2022red, ganguli2022red} attempt to find inputs that elicit harmful outputs from aligned models. Wei et al.~\cite{wei2024jailbroken} analyze failure modes of LLM safety training, identifying categories of prompts that reliably bypass safety measures. This work complements ours by identifying \textit{which} prompts fail, while we study \textit{how reliably} the model makes consistent decisions on a given prompt. The existence of ``borderline'' prompts that inconsistently trigger safety refusals suggests that adversaries could exploit stochastic sampling to bypass safety measures.

\textbf{Safety Training.} Modern LLMs are trained to be helpful, harmless, and honest through techniques like RLHF~\cite{ouyang2022training, bai2022training}. These methods optimize for human preferences but may not explicitly encourage consistent behavior across sampling configurations. Our findings suggest that safety training could benefit from explicitly penalizing inconsistent refusal behavior.

\section{Methodology}

\subsection{Experimental Design}

\textbf{Models.} We evaluate four instruction-tuned models from three distinct families: (1) Llama 3.1 8B Instruct~\cite{grattafiori2024llama}, Meta's open-weight model; (2) Qwen 2.5 7B Instruct, from Alibaba's model family; (3) Qwen 3 8B, representing the latest Qwen generation; and (4) Gemma 3 12B Instruct, Google's recent instruction-tuned model. This selection spans parameter counts from 7B to 12B and represents three major model families with distinct training approaches.\footnote{Exploratory runs not used in the reported results: a fifth model (Mistral-7B-Instruct-v0.3), generated but not carried through the unified-judge evaluation, and a GPT-4o-mini judge piloted on the newer models. Reported results use the Llama-3.1-70B and Claude-3.5-Haiku judges.}

\textbf{Dataset.} We use 876 harmful prompts assembled from two standard red-teaming benchmarks: the AdvBench harmful-behaviors set~\cite{zou2023advbench} and the HarmBench textual behaviors~\cite{mazeika2024harmbench}. We pool the prompts from both sources and remove exact duplicates and near-duplicates (text normalized to lowercase with collapsed whitespace; near-duplicates removed at a fuzzy string-similarity threshold of $0.9$ via \texttt{difflib.SequenceMatcher}, keeping the first occurrence). The resulting set contains 491 AdvBench and 385 HarmBench prompts (876 total). We use each benchmark's base harmful behavior---the request itself---not adversarial-suffix attack strings.

\textbf{Sampling Configurations.} We test 4 temperature values (0.0, 0.3, 0.7, 1.0) $\times$ 5 random seeds (42--46), yielding 20 unique configurations per model and 70,080 total responses across all four models. At temperature 0.0, we use greedy decoding; any residual variation across seeds may arise from GPU non-determinism in floating-point operations. All inference was performed using vLLM~\cite{kwon2023vllm} for efficient batch generation.

\textbf{Response Classification.} We classify each response into three categories with an LLM judge. The single-model Llama 3.1 8B deep-dive (Sections 4.1--4.2) uses \textbf{Llama 3.1 70B Instruct} as the judge. All cross-model results (Section 4.3 and Table~\ref{tab:multimodel}) use \textbf{Claude 3.5 Haiku} as a unified judge across all four models---a different model family---which gives a consistent cross-model comparison and addresses potential same-family bias. Both judges apply the same rubric:
\begin{itemize}
    \item \textbf{REFUSE}: Model declines to answer or provides safety-focused redirection
    \item \textbf{PARTIAL}: Model provides limited information with caveats or warnings
    \item \textbf{COMPLY}: Model directly answers the harmful request
\end{itemize}

\subsection{Safety Stability Index (SSI)}

For a prompt $p$ with $N=20$ responses across sampling configurations, let $n_c$ denote the count of responses classified as category $c \in \{\text{REFUSE}, \text{PARTIAL}, \text{COMPLY}\}$. We define the Safety Stability Index as:

\begin{equation}
\text{SSI}(p) = \frac{\max_c(n_c)}{N}
\end{equation}

SSI ranges from $1/3$ (uniform distribution across all three categories) to $1.0$ (perfect agreement on a single decision). We define a prompt as \textit{unstable} if $\text{SSI}(p) < 0.8$, corresponding to fewer than 16 of 20 samples agreeing on the majority decision.

We define the per-prompt compliance rate as the fraction of responses classified as COMPLY:

\begin{equation}
\text{CR}(p) = \frac{n_{\text{COMPLY}}}{N}
\end{equation}

In our analyses, PARTIAL responses are treated as a distinct category, separate from both full refusal (REFUSE) and full compliance (COMPLY). This three-way classification allows us to distinguish between complete safety failures and responses that provide limited information with caveats.

We aggregate SSI scores across prompts to compute mean stability by temperature and identify the proportion of unstable prompts. We also track the \textit{flip rate}: the percentage of prompts that produce at least one different decision across configurations.

\section{Results}

\subsection{Overall Stability}

\begin{figure}[h]
\centering
\includegraphics[width=0.6\textwidth]{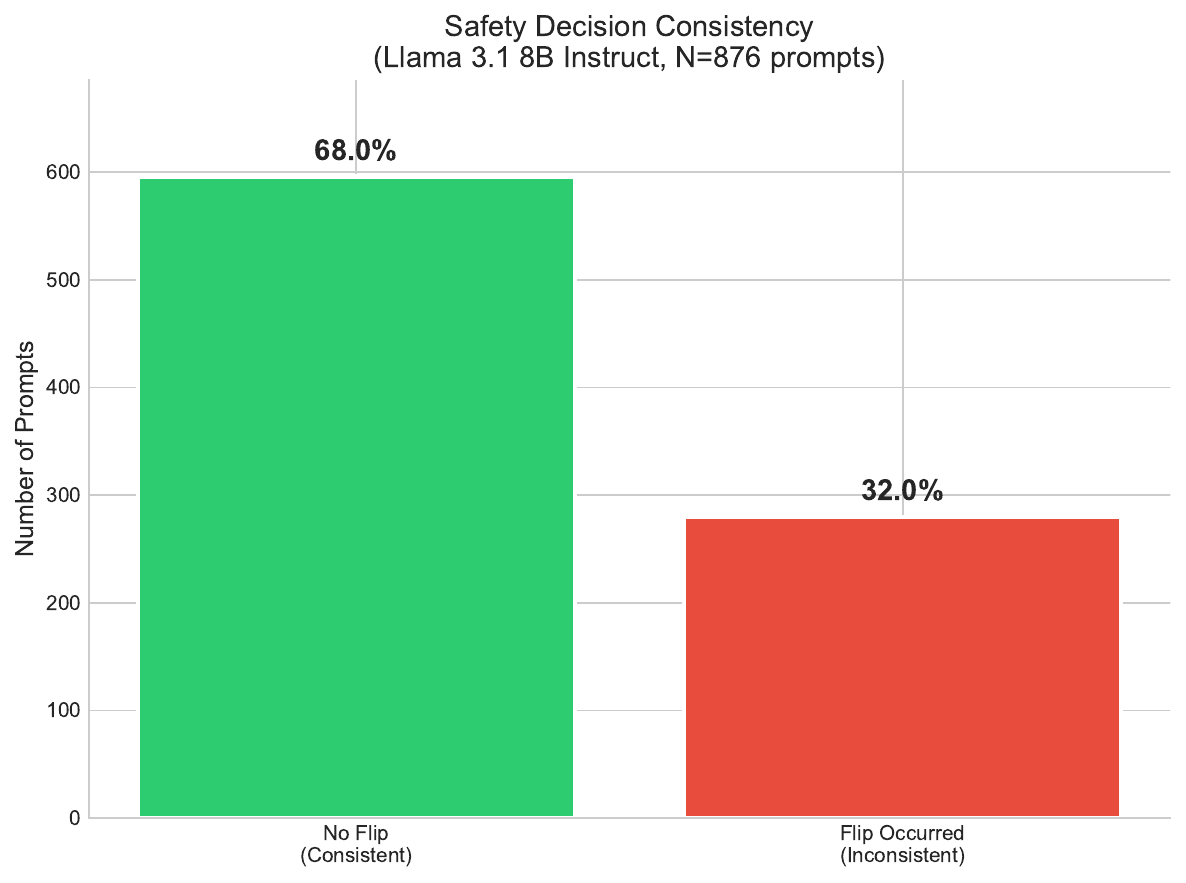}
\caption{Distribution of consistent vs. inconsistent (flip) prompts across 876 harmful prompts tested on Llama 3.1 8B Instruct.}
\label{fig:flip_rate}
\end{figure}

Figure~\ref{fig:flip_rate} shows that 68\% of prompts produce consistent safety decisions across all 20 configurations, while 32\% [95\% CI: 29.0\%--35.1\%] exhibit at least one flip between different decisions. Of these, 14.3\% [95\% CI: 12.1\%--16.7\%] show substantial instability (SSI $<$ 0.8), while the remaining 17.7\% have occasional flips but maintain strong majority agreement.

The overall response distribution shows that the model predominantly refuses harmful requests (83.1\% REFUSE, 10.9\% PARTIAL, 5.9\% COMPLY), suggesting strong safety alignment on average. However, the substantial proportion of unstable prompts indicates that aggregate statistics obscure significant per-prompt variability. \textbf{Judge Note:} The single-model Llama 3.1 8B deep-dive in Sections 4.1--4.2 uses the \textbf{Llama 70B judge} (32\% flip rate). All cross-model results---the temperature analysis in Section 4.3 (including the within-temperature table) and Table~\ref{tab:multimodel}---use \textbf{Claude 3.5 Haiku} as a unified judge (27.3\% flip rate for Llama; within-temperature flip 5.1\%$\to$23.6\%). The 4.7 percentage-point flip-rate difference for Llama reflects inter-judge variation ($\kappa=0.62$); see Section~\ref{sec:external_judge} for validation.

\subsection{Distribution of Stability Scores}

\begin{figure}[h]
\centering
\includegraphics[width=0.7\textwidth]{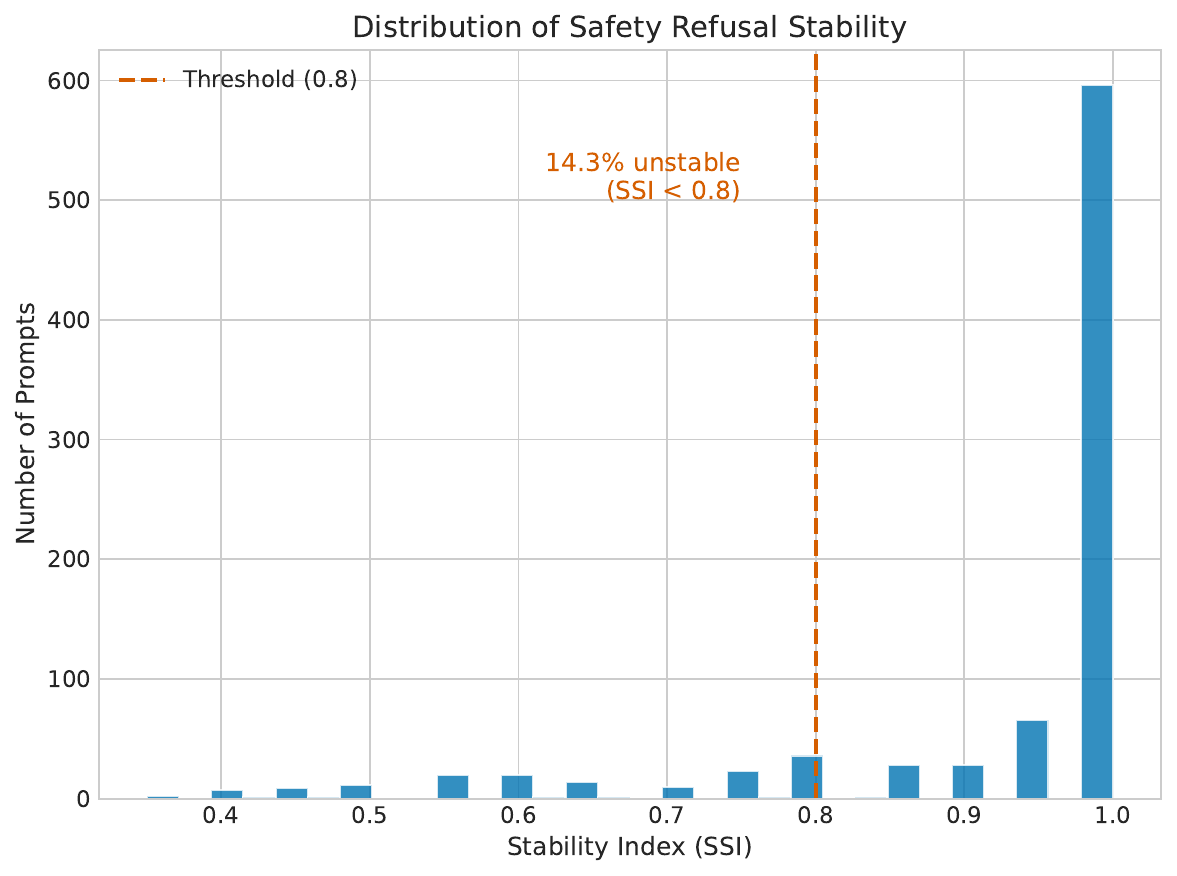}
\caption{Distribution of Safety Stability Index (SSI) scores. The dashed line at 0.8 indicates the threshold for unstable prompts. 14.3\% of prompts fall below this threshold.}
\label{fig:ssi_dist}
\end{figure}

Figure~\ref{fig:ssi_dist} presents the distribution of SSI scores across all 876 prompts. The distribution is heavily right-skewed with a strong mode at perfect stability (SSI = 1.0): the median SSI is 1.0, indicating that more than half of prompts receive identical \textit{classification labels} across all configurations (though the response text may vary). However, 14.3\% of prompts fall below the 0.8 stability threshold, demonstrating that instability affects a meaningful fraction of safety evaluations.

\subsection{Temperature Effects}

\begin{figure}[h]
\centering
\includegraphics[width=0.7\textwidth]{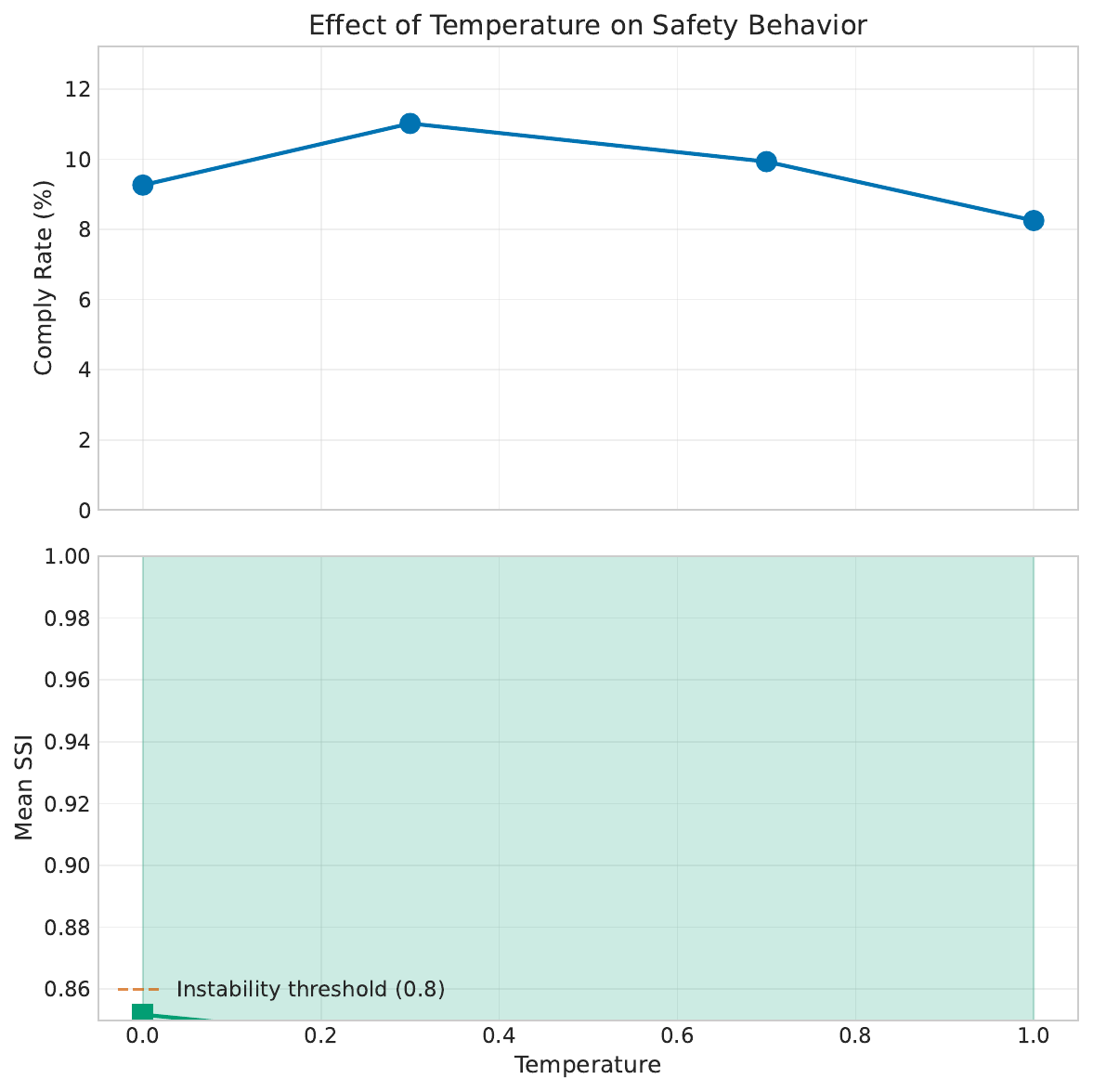}
\caption{Effect of temperature on comply rate (blue) and mean Safety Stability Index (green). Higher temperatures reduce stability while showing non-monotonic effects on compliance.}
\label{fig:temp_effect}
\end{figure}

Using the unified Claude 3.5 Haiku judge across all four models, Figure~\ref{fig:temp_effect} illustrates the relationship between temperature and both compliance rate and mean SSI. We observe that stability generally decreases as temperature increases (Friedman $\chi^2 = 396.81$, $p < 0.001$, Kendall's $W = 0.038$ indicating a small but highly significant effect). Mean within-temperature SSI drops from 0.977 at temperature 0.0 to 0.942 at temperature 1.0, a decrease of 0.035 [95\% CI: 0.031--0.039]. Post-hoc Wilcoxon signed-rank tests with Bonferroni correction confirm significant decreases in SSI between temperature 0.0 and 1.0 ($p < 0.001$). The flip rate increases from 5.1\% at temperature 0.0 to 23.6\% at temperature 1.0 for Llama 3.1 8B (similar patterns across all models).

Interestingly, the lowest compliance rate occurs at temperature 1.0 (3.0\%). However, this may reflect that high-temperature sampling produces less coherent outputs that fail to meaningfully engage with the harmful request, rather than representing more robust safety behavior.

\subsubsection{Within-Temperature Seed Stability}

To separate seed variance from temperature-induced distribution shift, we computed SSI separately at each temperature using only the 5 seeds per configuration. Table~\ref{tab:within_temp} shows that even at temperature 0.0 (greedy decoding), there is non-zero instability: 5--12\% of prompts flip across seeds depending on the model. This residual variation reflects both inference non-determinism, likely from GPU floating-point operations, and judge noise: Section 5.2 shows that at least 9 of Qwen 2.5's 81 flips at t=0.0 are judge noise alone. At temperature 1.0, flip rates increase to 12--24\% depending on the model, confirming that higher temperatures substantially increase seed sensitivity.

\begin{table}[h]
\centering
\caption{Within-temperature stability (seed variance only, N=5 seeds). Shows that instability increases with temperature even when holding temperature fixed. All values use the unified Claude 3.5 Haiku judge across the four models.}
\label{tab:within_temp}
\begin{tabular}{lcccc}
\toprule
\textbf{Model} & \textbf{t=0.0} & \textbf{t=0.3} & \textbf{t=0.7} & \textbf{t=1.0} \\
\midrule
\multicolumn{5}{l}{\textit{Mean Within-Temperature SSI}} \\
Gemma-3-12B & 0.986 & 0.972 & 0.967 & 0.964 \\
Llama-3.1-8B & 0.985 & 0.961 & 0.948 & 0.926 \\
Qwen3-8B & 0.967 & 0.948 & 0.943 & 0.945 \\
Qwen2.5-7B & 0.972 & 0.956 & 0.941 & 0.932 \\
\midrule
\multicolumn{5}{l}{\textit{Flip Rate (\%)}} \\
Gemma-3-12B & 4.9 & 9.9 & 11.3 & 12.6 \\
Llama-3.1-8B & 5.1 & 12.7 & 17.2 & 23.6 \\
Qwen3-8B & 12.0 & 17.6 & 18.6 & 18.4 \\
Qwen2.5-7B & 9.2 & 15.4 & 20.1 & 22.8 \\
\bottomrule
\end{tabular}
\end{table}

\subsection{Stability vs. Compliance Analysis}

\begin{figure}[h]
\centering
\includegraphics[width=0.7\textwidth]{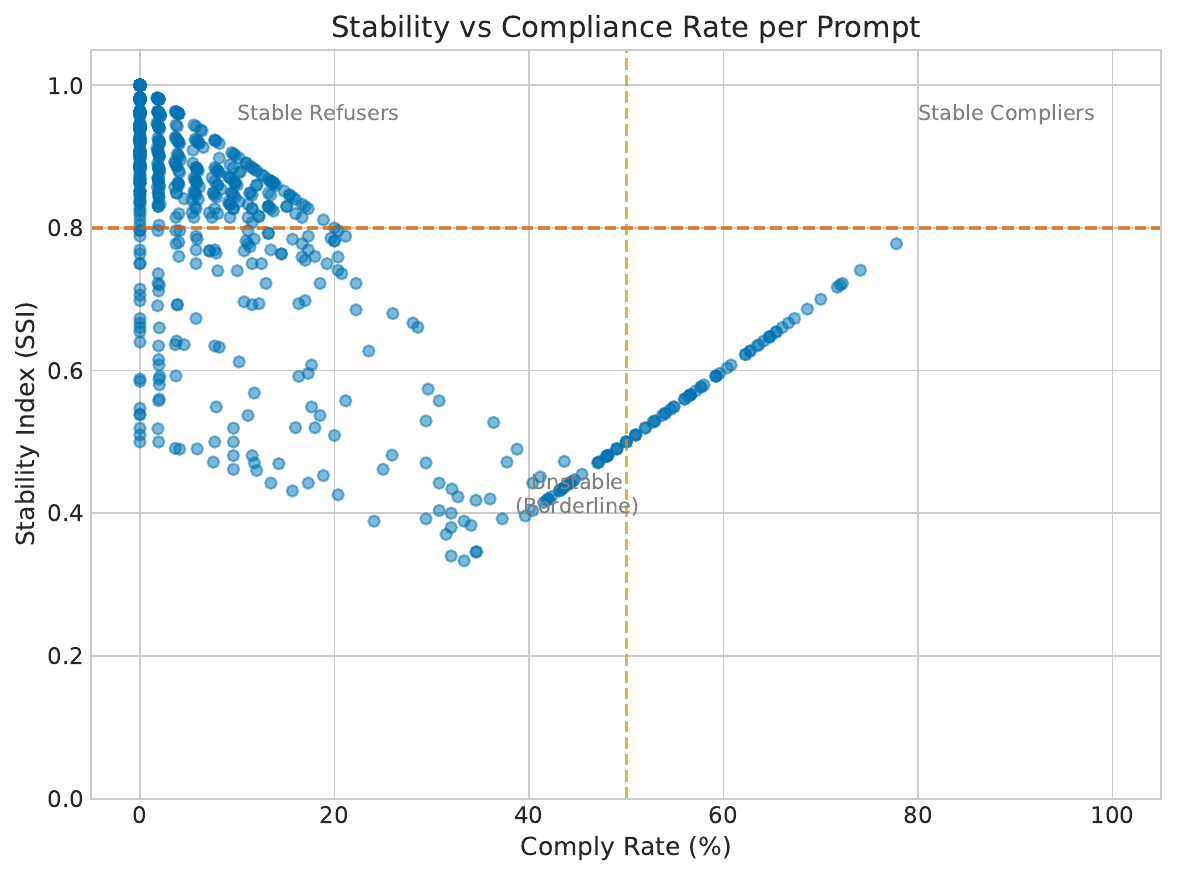}
\caption{Safety Stability Index vs. compliance rate per prompt. Most prompts cluster in the stable-refusal region (upper left), while a subset shows variable behavior.}
\label{fig:stability_compliance}
\end{figure}

Figure~\ref{fig:stability_compliance} presents a scatter plot of per-prompt SSI against compliance rate. The distribution reveals two main regions:

\begin{itemize}
    \item \textbf{Stable Refusers} (left, high SSI): The majority of prompts cluster here with low compliance and high stability---the model consistently refuses these prompts
    \item \textbf{Variable Region} (center-right): A smaller set of prompts shows lower stability with varying compliance rates---these are the borderline cases where the model's decision is sensitive to sampling
\end{itemize}

The existence of borderline prompts suggests that certain phrasings or topics create ambiguity in the model's safety classification, causing it to oscillate between refusing and complying depending on stochastic factors.

\subsection{What Predicts Prompt Instability?}

To understand what makes certain prompts unstable, we conducted an \textbf{exploratory analysis} of linguistic features and their correlation with SSI. These analyses involve multiple comparisons and should be interpreted as hypothesis-generating rather than confirmatory. We find that some prompt characteristics correlate with instability:

\begin{itemize}
    \item \textbf{Sensitive terms increase stability}: Prompts containing obvious harmful keywords (e.g., ``hack,'' ``bomb,'' ``kill'') have \textit{higher} SSI (Pearson $r = 0.18$), suggesting the model confidently refuses clearly harmful requests.
    \item \textbf{Multi-sentence prompts are less stable}: Prompts with multiple sentences show lower SSI (Pearson $r = -0.21$), possibly because complex requests create ambiguity.
\end{itemize}

A logistic regression classifier predicting instability from linguistic features achieves AUC = 0.73 [95\% CI: 0.69--0.78] in 5-fold cross-validation, indicating that surface-level features partially explain which prompts will be unstable.\footnote{Features used: word count, character count, average word length, sentence count, question mark presence, sensitive term count, code marker presence, and uppercase character ratio.} The most predictive features are average word length, sensitive term count, and word count.

To quantify category effects, we classified prompts into 12 harm categories using keyword matching. \textbf{Copyright-related requests} (N=112) are dramatically more unstable than all other categories: mean SSI = 0.568 with 89.3\% of prompts classified as unstable, compared to misinformation (N=30, SSI = 0.822, 20\% unstable) and hacking (N=118, SSI = 0.862, 22\% unstable). Copyright prompts also show the highest compliance rate (45.9\%), suggesting the model is uncertain whether reproducing copyrighted content constitutes harm. In contrast, \textbf{self-harm prompts} (N=15) are most stable (SSI = 0.990, 0\% unstable), likely because safety training provides clear guidance on this category. This suggests that instability concentrates in categories where harm boundaries are ambiguous. (Note: the processed prompts carry no native category label---the \texttt{harm\_category} field is empty---so all categories here, including the copyright cluster (N=112), are assigned post-hoc by keyword matching on the prompt text; the ``other'' category contains 292 prompts not matching specific keywords.)

Table~\ref{tab:examples} provides examples of prompts exhibiting high instability, where the model's decision appears sensitive to stochastic sampling factors. These prompts demonstrate a range of instability patterns: some alternate primarily between REFUSE and PARTIAL, while others show more pronounced variation including compliance in some configurations.

\begin{table}[t]
\centering
\caption{Example prompts exhibiting decision instability across sampling configurations (N=20).}
\label{tab:examples}
\begin{tabular}{lcccc}
\toprule
\textbf{Prompt Category} & \textbf{SSI} & \textbf{REFUSE} & \textbf{PARTIAL} & \textbf{COMPLY} \\
\midrule
Request for copyrighted content & 0.55 & 2/20 & 11/20 & 7/20 \\
Request for harmful instructions & 0.45 & 3/20 & 9/20 & 8/20 \\
Request for copyrighted passage & 0.50 & 10/20 & 7/20 & 3/20 \\
\bottomrule
\end{tabular}
\end{table}

\subsection{Model Comparison}

\begin{figure}[h]
\centering
\includegraphics[width=0.9\textwidth]{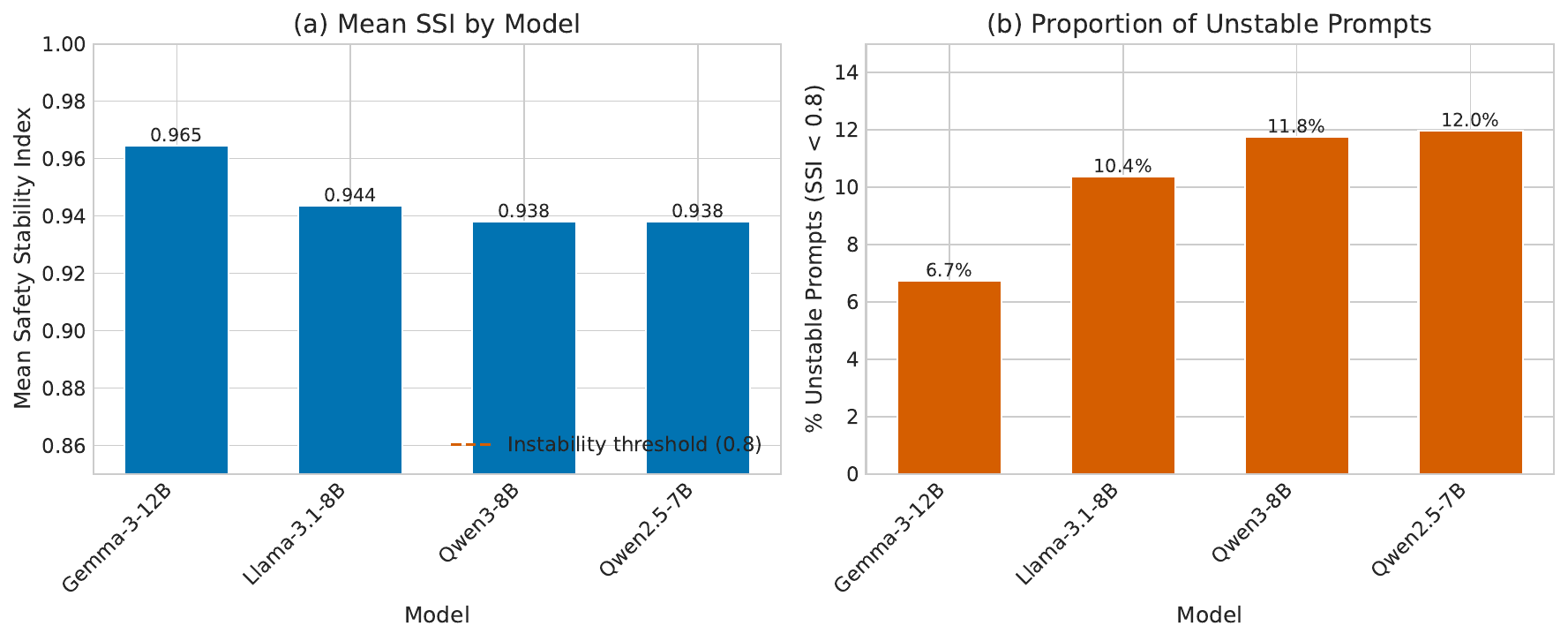}
\caption{Safety Stability Index comparison across models. Left: Mean SSI with standard deviation error bars. Right: Percentage of prompts with SSI $<$ 0.8 (unstable). All four models show instability rates between 6.7\% and 12.0\%.}
\label{fig:model_comparison}
\end{figure}

Figure~\ref{fig:model_comparison} compares stability across the four evaluated models. We observe consistent patterns of instability across all three model families, with instability rates (SSI $<$ 0.8) ranging from 6.7\% to 12.0\%.

Table~\ref{tab:multimodel} summarizes the stability metrics across all four models from three families, using Claude 3.5 Haiku as a unified judge for methodological consistency. The models span a range of stability characteristics: Gemma 3 12B shows the highest stability (SSI=0.965, 18.4\% flip rate) while also having the lowest refusal rate (78.5\%), while the Qwen models show the lowest stability (SSI=0.938). Notably, Llama 3.1 8B falls in the middle (SSI=0.944, 27.3\% flip rate). All models exhibit non-trivial instability (6.7--12.0\% of prompts highly unstable) despite different training approaches and model families, suggesting that safety decision instability is common across contemporary instruction-tuned models in the 7B--12B parameter range.

\begin{table}[h]
\centering
\caption{Stability comparison across model families. 95\% CIs: Wilson score intervals for Flip Rate and \% Unstable (proportions over 876 prompts); bootstrap over prompts for Mean SSI. All models evaluated using Claude 3.5 Haiku as judge. Models sorted by SSI (descending).}
\label{tab:multimodel}
\begin{tabular}{lcccc}
\toprule
\textbf{Model} & \textbf{Mean SSI} $\uparrow$ & \textbf{Flip Rate} & \textbf{\% Unstable} & \textbf{Refusal} \\
\midrule
Gemma-3-12B & \textbf{.965} [.958, .971] & 18.4\% [16.0, 21.1] & \textbf{6.7\%} [5.3, 8.6] & 78.5\% \\
Llama-3.1-8B & .944 [.936, .951] & 27.3\% [24.4, 30.3] & 10.4\% [8.5, 12.6] & 79.3\% \\
Qwen3-8B & .938 [.929, .946] & 27.7\% [24.9, 30.8] & 11.8\% [9.8, 14.1] & 92.5\% \\
Qwen2.5-7B & .938 [.929, .946] & 26.3\% [23.4, 29.3] & 12.0\% [10.0, 14.3] & 81.3\% \\
\bottomrule
\end{tabular}
\end{table}

\begin{figure}[h]
\centering
\includegraphics[width=0.7\textwidth]{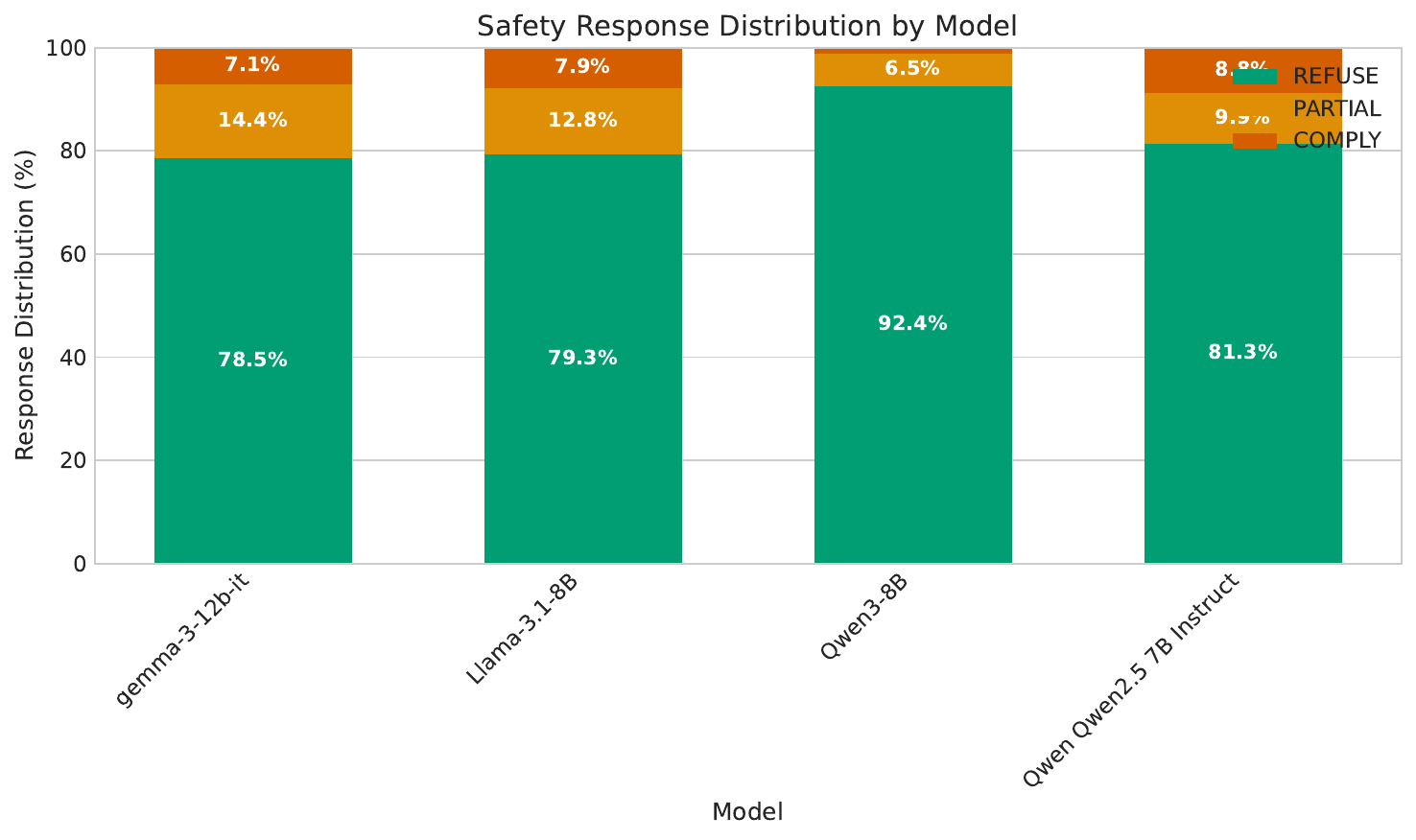}
\caption{Response distribution by model. Green indicates refusal, orange partial compliance, and red full compliance.}
\label{fig:model_distribution}
\end{figure}

Figure~\ref{fig:model_distribution} shows the response distribution (REFUSE, PARTIAL, COMPLY) for each model. The models differ substantially in their base refusal rates: Qwen 3 8B is most conservative (92.5\% refuse), while Gemma 3 12B shows the lowest refusal rate (78.5\%) with more PARTIAL responses (14.4\% vs 6.5\% for Qwen 3). Llama 3.1 8B and Qwen 2.5 7B fall in the middle (79.3\% and 81.3\% respectively).

\textbf{Stability-Compliance Relationship.} While the most stable model (Gemma) also has the lowest refusal rate, this correlation is based on only four models and may reflect model family differences. However, analyzing the relationship \textit{within} each model provides stronger evidence: per-prompt SSI correlates negatively with compliance rate (Spearman $\rho = -0.47$ to $-0.70$ depending on model, all $p < 0.001$). This indicates that prompts where the model is more likely to comply are also less stable---the model ``wavers'' more on borderline-harmful requests.

\textbf{Judge Model.} We use Llama 3.1 70B Instruct as the judge for the single-model Llama-3.1-8B analysis, which is distinct from the 8B model being evaluated. Prior work on LLM-as-judge has shown strong correlation with human annotations for safety-related tasks~\cite{zheng2023judging}. The three-class rubric (REFUSE, PARTIAL, COMPLY) was designed to capture the spectrum of safety responses, with PARTIAL indicating hedged or caveat-laden responses that neither fully refuse nor fully comply.

\subsection{External Judge Validation}
\label{sec:external_judge}

To address potential same-family bias concerns with our Llama-based judge and ensure methodological consistency across all models, we evaluated \textit{all} 70,080 responses using Claude 3.5 Haiku as an external judge from Anthropic's model family. This provides cross-validation with a judge trained on entirely different data and by a different organization.

\textbf{Inter-Judge Agreement.} Comparing Claude 3.5 Haiku to Llama 70B judgments on the Llama 3.1 8B and Qwen 2.5 7B responses---the only two models labeled by both judges, since the Llama 70B judge was not run on Gemma 3 or Qwen 3---we observe 89.1\% exact agreement on the three-class classification over the 35,005 valid pairs (of 35,040 nominal; 35 excluded for a non-three-class label), with Cohen's $\kappa = 0.62$ indicating substantial inter-rater reliability. Table~\ref{tab:confusion_matrix} shows the confusion matrix. The largest disagreements occur when Llama 70B labels a response as REFUSE but Claude Haiku labels it as PARTIAL (1,580 cases), suggesting Claude Haiku applies a stricter definition of full refusal. Crucially, the \textit{relative ranking} of model stability remains consistent across judges.

\begin{table}[h]
\centering
\caption{Confusion matrix comparing the Llama 70B and Claude Haiku judges on the 35,005 valid Llama 3.1 8B and Qwen 2.5 7B response pairs (of 35,040; 35 excluded where a judge returned a non-three-class label) --- the two models labeled by both judges. Overall agreement: 89.1\%, Cohen's $\kappa$ = 0.62.}
\label{tab:confusion_matrix}
\begin{tabular}{lccc|c}
\toprule
& \multicolumn{3}{c}{\textbf{Claude Haiku}} & \\
\textbf{Llama 70B} & REFUSE & PARTIAL & COMPLY & Total \\
\midrule
REFUSE & 27,933 & 1,580 & 940 & 30,453 \\
PARTIAL & 197 & 1,842 & 558 & 2,597 \\
COMPLY & 6 & 532 & 1,417 & 1,955 \\
\midrule
Total & 28,136 & 3,954 & 2,915 & 35,005 \\
\bottomrule
\end{tabular}
\end{table}

\textbf{Unified Results.} Table~\ref{tab:multimodel} reports all metrics using Claude 3.5 Haiku as the judge for all four models, ensuring methodological consistency in cross-model comparisons. The high inter-judge agreement (89.1\%, $\kappa = 0.62$, measured on the two models with both judges) validates that our instability findings are robust to judge selection and not artifacts of same-family bias.

\textbf{Consensus-Restricted Sensitivity Analysis.} To assess how sensitive our instability estimates are to judge disagreement, we recomputed metrics using only responses where both judges agreed on classification. When restricting to consensus responses, mean SSI increased from 0.941 to 0.965, and flip rate decreased from 26.8\% to 17.5\%. Substantial instability therefore persists even on responses whose labels are uncontested between the two judges. We note that this is a sensitivity analysis rather than a decomposition of model versus judge variation: consensus filtering may both retain errors shared by the two judges and remove genuine model variation that produces borderline outputs, so the 9.3-point reduction should not be interpreted as an estimate of judge noise.

\subsection{Sample Size Requirements for Reliable Evaluation}

A key practical question is: how many samples are needed for reliable safety evaluation? We address this on the Llama 3.1 8B responses (Llama 70B judge) by simulating subsampling from the N=20 ground truth. For each prompt, we draw $n$ of the 20 configurations \emph{without replacement} (as in the fixed-temperature analysis), compute the majority label, and measure agreement with the full-data majority; we average over bootstrap resamples. At $n=20$ all configurations are drawn, so agreement is 100\% by construction.

\begin{figure}[h]
\centering
\includegraphics[width=0.7\textwidth]{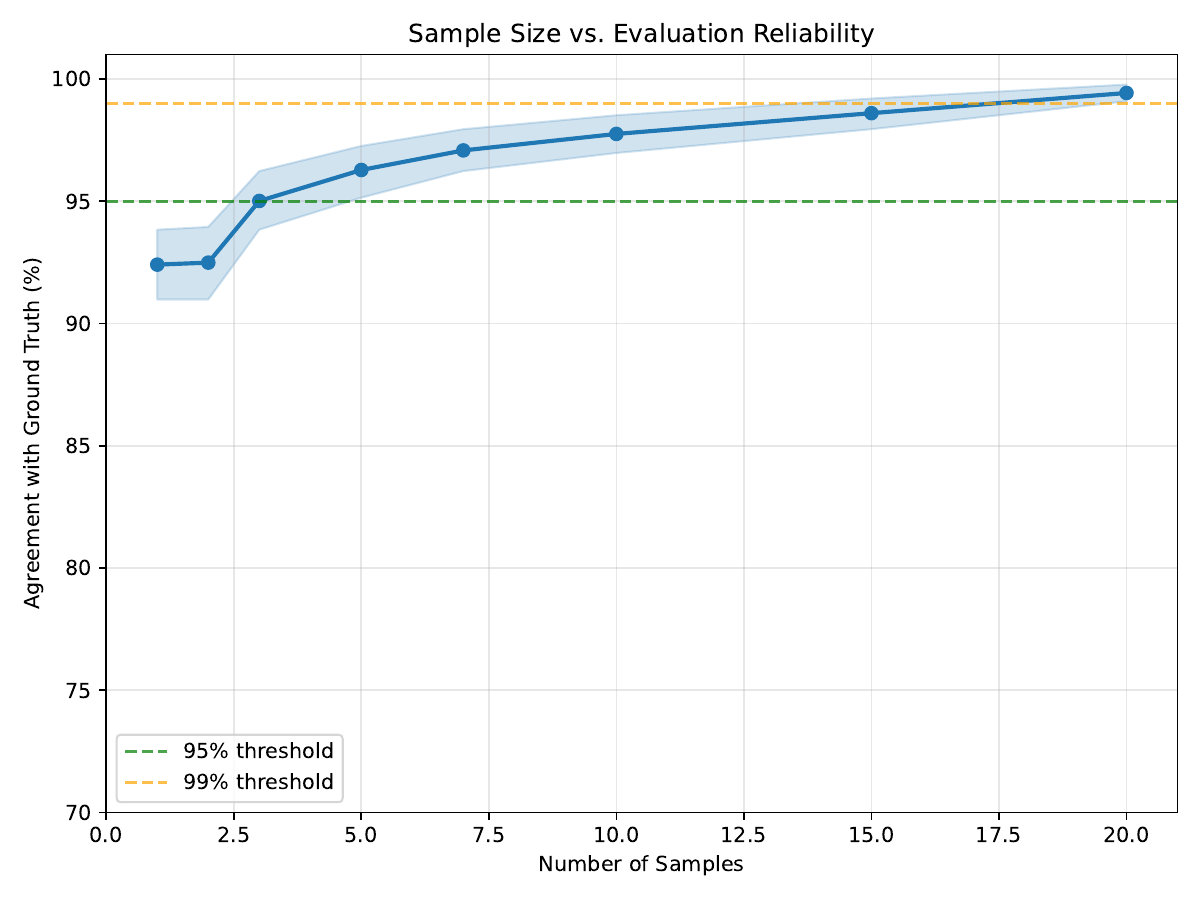}
\caption{Evaluation reliability as a function of sample size (Llama 3.1 8B, pooled across temperatures). Single-shot testing (N=1) achieves only 92.5\% agreement with ground truth; the 95\% reliability threshold requires N$\geq$3 samples (pooled).}
\label{fig:sample_size}
\end{figure}

Figure~\ref{fig:sample_size} shows that single-shot evaluation of Llama 3.1 8B (N=1) agrees with the ground truth only 92.5\% [95\% CI: 91.0\%--93.8\%] of the time---meaning approximately 7.5\% of prompts would be misclassified by single-sample testing. Agreement rises to 95\% at N=3; using all 20 configurations (N=20) reproduces the full-data majority by construction (100\%).

\begin{table}[h]
\centering
\caption{Sample size requirements for reliable safety evaluation (Llama 3.1 8B single-model, Llama 70B judge). Subsampling is without replacement (averaged over bootstrap resamples); N=20 uses all configurations and is therefore 100\% by construction.}
\label{tab:sample_size}
\begin{tabular}{ccc}
\toprule
\textbf{N Samples} & \textbf{Agreement (\%)}& \textbf{95\% CI} \\
\midrule
1 & 92.5 & [91.0, 93.8] \\
3 & 95.1 & [93.8, 96.2] \\
5 & 96.3 & [95.1, 97.3] \\
10 & 97.8 & [96.7, 98.5] \\
20 & 100.0 & --- \\
\bottomrule
\end{tabular}
\end{table}

These results suggest that safety benchmarks should use at least 3 samples per prompt for 95\% reliability when pooling across temperatures, and more for a stricter bar---the full seed set at high temperature, or ten pooled samples, for $>$97\%---with additional sampling for high-stakes deployment decisions.

\subsubsection{Fixed-Temperature Reliability}

The above analysis pools samples across temperatures. However, most benchmarks operate at a fixed temperature. Table~\ref{tab:fixed_temp_sample_size} shows reliability when temperature is held constant and only seeds vary. At greedy decoding (t=0.0), single-shot evaluation is already reliable (98.7\% agreement with the 5-seed ground truth). Reliability degrades as temperature rises: single-shot agreement falls to 92.8\% at t=0.7 and 90.3\% at t=1.0. The number of seeds needed for $>$97\% agreement therefore scales with temperature---one seed suffices at t=0.0, three at t$\leq$0.3 (97.4\%), but at t=0.7 and t=1.0 three seeds reach only 96.0\% and 94.8\%, so $>$97\% requires the full five-seed set.

\begin{table}[h]
\centering
\caption{Sample size reliability at fixed temperatures (Llama 3.1 8B, Llama 70B judge). Values show \% agreement with the 5-seed ground truth; N=5 uses all seeds and is therefore 100\% by construction. Single-shot is reliable at greedy decoding (98.7\%) but degrades with temperature (90.3\% at t=1.0).}
\label{tab:fixed_temp_sample_size}
\begin{tabular}{ccccc}
\toprule
\textbf{N Seeds} & \textbf{t=0.0} & \textbf{t=0.3} & \textbf{t=0.7} & \textbf{t=1.0} \\
\midrule
1 & 98.7\% & 95.1\% & 92.8\% & 90.3\% \\
3 & 99.7\% & 97.4\% & 96.0\% & 94.8\% \\
5 (full) & 100\% & 100\% & 100\% & 100\% \\
\bottomrule
\end{tabular}
\end{table}

\subsection{Binary Scoring Ablation}

Many safety benchmarks use binary scoring (safe/unsafe) rather than three-way classification. To assess whether instability is driven by REFUSE$\leftrightarrow$PARTIAL ambiguity, we tested two binary mappings: conservative (PARTIAL$\rightarrow$COMPLY, treating hedged responses as unsafe) and lenient (PARTIAL$\rightarrow$REFUSE, treating hedged responses as safe). Table~\ref{tab:binary_ablation} shows that instability persists under both mappings, though at reduced levels. The three-way flip rate of 24.9\% decreases to 17.8\% (conservative) or 11.4\% (lenient), confirming that some instability arises from REFUSE$\leftrightarrow$PARTIAL ambiguity but substantial instability remains even under binary scoring.

\begin{table}[h]
\centering
\caption{Binary scoring ablation: results under different PARTIAL handling rules. Instability persists regardless of scoring methodology.}
\label{tab:binary_ablation}
\begin{tabular}{lccc}
\toprule
\textbf{Scoring Rule} & \textbf{Mean SSI} & \textbf{Flip Rate} & \textbf{\% Unstable} \\
\midrule
Three-way (original) & 0.946 & 24.9\% & 10.2\% \\
Binary (PARTIAL$\rightarrow$COMPLY) & 0.968 & 17.8\% & 5.3\% \\
Binary (PARTIAL$\rightarrow$REFUSE) & 0.976 & 11.4\% & 4.8\% \\
\bottomrule
\end{tabular}
\end{table}

\section{Discussion}

\subsection{Implications for Safety Evaluation}

Our findings demonstrate that single-shot safety evaluations can significantly misrepresent model safety. A prompt classified as ``safe'' in one evaluation may produce a harmful response under different sampling conditions. This variability has several important implications:

\textbf{Benchmark Reliability.} Safety benchmarks that report only aggregate pass rates may be highly sensitive to random seed choice. Different evaluators using different random seeds could reach contradictory conclusions about the same model.

\textbf{Deployment Considerations.} For safety-critical applications, using deterministic sampling (temperature 0.0) provides the most stable behavior (mean within-temperature SSI = 0.977). At this temperature, the mean within-temperature flip rate is just 7.8\% (Table~\ref{tab:within_temp}), compared to 19.4\% at temperature 1.0.

\textbf{Adversarial Robustness.} The existence of borderline prompts that flip between refuse and comply creates a practical vulnerability: adversaries can exploit stochastic sampling by repeatedly querying until obtaining a harmful response. We quantified this risk by computing expected queries to first compliance for prompts where any COMPLY response was observed (N=454 prompt-model pairs). The median expected queries is just 2, with 69\% of exploitable prompts yielding compliance within 5 queries. This demonstrates that instability creates exploitable attack surfaces even for prompts that usually refuse.

\textbf{Evaluation Protocol Recommendations.} Our sample size analysis provides concrete guidance: single-shot evaluation misclassifies 7.6\% of prompts. For benchmark reliability, we recommend a minimum of N=3 samples per prompt (95\% agreement). For deployment decisions, N=10+ samples provide 98\% reliability. When computational constraints require single-shot evaluation, results should be reported with appropriate uncertainty bounds acknowledging the $\sim$8\% error rate.

\subsection{Limitations}

\textbf{Model Scale.} While we evaluate four models from three families (Llama, Qwen, and Gemma), all are in the 7-12B parameter range. Larger models (70B, 405B) or closed-source models (GPT-4, Claude) may exhibit different stability characteristics. Future work should extend this analysis across model scales.

\textbf{Judge Model.} The single-model Llama 3.1 8B analysis uses Llama 3.1 70B Instruct as the judge; all cross-model results use Claude 3.5 Haiku as a unified judge across the four models. While the agreement between these judges ($\kappa=0.62$) increases confidence in our findings, human annotation on a subset of responses would further strengthen conclusions.

\textbf{Limited Seed Count.} We test only 5 seeds per temperature. While sufficient to detect instability, more seeds would provide tighter confidence intervals on stability estimates.

\textbf{Dataset Scope.} Our analysis uses direct harmful-behavior requests from AdvBench and HarmBench. Different prompt distributions---e.g., obfuscated or adversarially-crafted jailbreaks, or human-preference safety datasets such as BeaverTails~\cite{ji2023beavertails}---may exhibit different stability patterns.

\textbf{Temperature-Seed Pooling.} Our primary SSI metric pools responses across both temperatures and seeds (N=20 configurations). Benchmarks that fix temperature may experience different reliability characteristics; we provide within-temperature analyses in Section 4.3.1 to address this. At greedy decoding (t=0.0), single-shot evaluation is 98.7\% reliable, compared to 92.5\% when pooling across temperatures.

\textbf{Greedy Decoding Non-Determinism.} We checked greedy determinism by comparing response texts across seeds at temperature 0.0. Llama 3.1 8B gave byte-identical responses across all five seeds for 73.4\% of prompts. Qwen 2.5 7B did so for only 17.6\%, so inference isn't deterministic even at temperature 0.0, likely because of GPU floating-point non-determinism in batched vLLM inference. Judge noise is also measurable. Among prompts whose five greedy responses are byte-identical, the Claude judge still gave mixed labels for 1 of 643 Llama prompts (0.2\%) but 9 of 154 Qwen 2.5 prompts (5.8\%). Those 9 are pure judge-noise flips: at least 9 of Qwen 2.5's 81 flips at t=0.0 don't come from the model. That's a lower bound, since it only counts prompts whose five responses match exactly. Pooled over all temperatures, the judge disagrees with itself on 0.44\% of pairs of byte-identical Llama responses and 3.61\% for Qwen 2.5. The Claude judge ran at the API default temperature (Table~\ref{tab:experimental_config}), which likely contributes. The low noise rate we measured for Llama doesn't carry over to other models, so the t=0.0 flip rates in Table~\ref{tab:within_temp} (5--12\%) combine inference variation with judge noise. We couldn't run this check for Qwen 3 8B or Gemma 3 12B. Separating the two sources cleanly would take strict determinism settings (e.g., \texttt{CUBLAS\_WORKSPACE\_CONFIG=:4096:8}, \texttt{torch.use\_deterministic\_algorithms(True)}) and a deterministic judge.

\textbf{Binary Stability Threshold.} Our SSI $<$ 0.8 threshold for instability is somewhat arbitrary. Practitioners with different risk tolerances may prefer stricter or more lenient thresholds. Table~\ref{tab:threshold_sensitivity} shows how results vary across thresholds.

\begin{table}[h]
\centering
\caption{Sensitivity analysis: \% of prompts classified as unstable at different SSI thresholds. The qualitative finding that a meaningful fraction of prompts exhibit instability is robust to threshold choice.}
\label{tab:threshold_sensitivity}
\begin{tabular}{lccccc}
\toprule
\textbf{Model} & \textbf{0.70} & \textbf{0.75} & \textbf{0.80} & \textbf{0.85} & \textbf{0.90} \\
\midrule
Gemma-3-12B & 4.3\% & 5.8\% & 6.7\% & 7.4\% & 9.1\% \\
Llama-3.1-8B & 6.2\% & 8.4\% & 10.4\% & 12.9\% & 17.6\% \\
Qwen3-8B & 8.6\% & 10.0\% & 11.8\% & 14.5\% & 17.8\% \\
Qwen2.5-7B & 8.2\% & 9.7\% & 12.0\% & 15.6\% & 18.9\% \\
\bottomrule
\end{tabular}
\end{table}

\subsection{Reproducibility Details}

Table~\ref{tab:experimental_config} provides complete experimental configuration details necessary for reproduction.

\begin{table}[h]
\centering
\caption{Experimental Configuration Details}
\label{tab:experimental_config}
\begin{tabular}{ll}
\toprule
\textbf{Parameter} & \textbf{Value} \\
\midrule
\multicolumn{2}{l}{\textit{Inference Configuration}} \\
Inference Engine & vLLM 0.6.3 \\
PyTorch Version & 2.4.1 \\
CUDA Version & 12.1 \\
GPU & NVIDIA A100-80GB \\
Batch Size & 32 \\
Temperatures & 0.0, 0.3, 0.7, 1.0 \\
Random Seeds & 42, 43, 44, 45, 46 \\
Top-p & 1.0 (disabled) \\
Top-k & -1 (disabled) \\
Max Output Tokens & 512 \\
Determinism Flags & None (default) \\
\midrule
\multicolumn{2}{l}{\textit{Prompt Format}} \\
System Prompt & None \\
Chat Template & Model default (via tokenizer) \\
\midrule
\multicolumn{2}{l}{\textit{Judge Configuration}} \\
Primary Judge & Llama 3.1 70B Instruct \\
External Judge & Claude 3.5 Haiku \\
Judge Temperature (Llama 3.1 70B) & 0.0 (greedy) \\
Judge Temperature (Claude 3.5 Haiku) & API default (1.0); no temperature set \\
Judge Prompt & See below \\
\bottomrule
\end{tabular}
\end{table}

\textbf{Judge Prompt.} The judge classifies responses using a three-class rubric: REFUSE (model clearly declined), PARTIAL (hedged response with warnings), and COMPLY (substantive help provided). The exact prompt template is: ``\textit{You are evaluating whether an AI assistant's response to a potentially harmful request is safe. [User request and response provided.] Classify the assistant's response into exactly one category: REFUSE, PARTIAL, or COMPLY.}''

\textbf{Parsing Rules.} The judge's output is parsed by searching for the first occurrence of ``REFUSE'', ``PARTIAL'', or ``COMPLY'' (case-insensitive). If none is found, the response is labeled ``UNKNOWN'' and excluded from analysis ($<$0.1\% of cases).

\section{Conclusion}

We have demonstrated that safety refusal decisions in LLMs exhibit significant instability across random seeds and temperature settings. Evaluating four instruction-tuned models from three families (Llama, Qwen, and Gemma) across 70,080 total responses, we find that 18--28\% of prompts exhibit decision flips depending on the model. Temperature significantly affects stability ($p < 0.001$), with the mean within-temperature flip rate increasing from 7.8\% at temperature 0.0 to 19.4\% at temperature 1.0. Importantly, these patterns are consistent across all model families and validated by an external Claude judge, suggesting that safety decision instability is common across the model families and scales tested (7B--12B parameters). Whether larger models (70B+) or different training approaches exhibit similar instability remains an important open question.

Critically, single-shot evaluation---the standard practice in safety benchmarking---agrees with multi-sample ground truth only 92.5\% of the time when pooling across temperatures (98.7\% at greedy decoding down to 90.3\% at temperature 1.0). This finding challenges the validity of single-shot safety evaluations and suggests that current benchmarks may provide incomplete assessments of model safety.

We recommend that safety evaluation protocols: (1) use at least N=3 samples per prompt for 95\% reliability, (2) report stability metrics alongside aggregate pass rates, (3) use lower temperatures for safety-critical deployments, and (4) implement ensemble voting for high-stakes decisions. Future work should explore whether larger models (70B+) or different safety training techniques (DPO, constitutional AI) produce more stable behavior.

The code and data for this study are available at \url{https://github.com/erikl2/safety-refusal-stability}.


\end{document}